\documentclass[runningheads]{llncs}
\usepackage[T1]{fontenc}
\usepackage{graphicx}  
\usepackage{xcolor}    

\usepackage{xcolor}

\definecolor{myRed}{RGB}{191,12,21}
\definecolor{myGreen}{RGB}{34, 139, 34}
\usepackage{amsmath}
\usepackage{booktabs}
\usepackage{multirow}
\usepackage{graphicx,verbatim}
\begin{document}
\title{Dino-NestedUNet : Unlocking Foundation Vision Encoders for Pathology Tumor Bulk Segmentation via Dense Decoding}
%

\author{Tianyang Wang\inst{1,2} \and
Ziyu Su\inst{1} \and
Abdul Rehman Akbar\inst{1,2} \and
Usama Sajjad\inst{1,2}
\and
Usman Afzaal\inst{1,2} 
\and
Lina Gokhale\inst{1} \and
Charles Rabolli\inst{1} \and
Wei Chen\inst{1} \and
Anil Parwani\inst{1} \and
M. Khalid Khan Niazi\inst{1,2}}
\authorrunning{T. Wang et al.}
\institute{
Department of Pathology, College of Medicine, The Ohio State University Wexner Medical Center, Columbus, OH, USA
\and
Department of Biomedical Engineering, The Ohio State University, Columbus, OH, USA\\
\email{Tianyang.Wang@osumc.edu}
}

\maketitle 
\begin{abstract}
Vision foundation models (VFMs), such as DINOv3, provide rich semantic representations that are promising for computational pathology. However, many current adaptations pair frozen VFMs with lightweight decoders, creating a capacity mismatch that often limits boundary fidelity for infiltrative tumor bulk segmentation. This paper presents \textbf{Dino-NestedUNet}, a framework that couples a pre-trained DINOv3 encoder with a Nested Dense Decoder. Instead of sparse skip connections and linear upsampling, the proposed decoder forms a dense grid of intermediate pathways to enable continuous feature reuse and multi-scale recalibration, aligning high-level semantics with low-level morphological textures during reconstruction. We evaluate Dino-NestedUNet on three histopathology cohorts (multi-center CHTN, institutional OSU, and CAMELYON16) and observe consistent improvements over UNet++ and standard Dino-UNet variants, particularly under cross-domain shift. To further assess external generalization, we perform zero-shot evaluation by training on CHTN and directly testing on unseen TIGER WSIBULK and OSU CRC cohorts without fine-tuning. These results suggest that dense decoding is a key ingredient for unlocking foundation encoders in boundary-sensitive pathology segmentation.

\keywords{Tumor segmentation \and Vision foundation models \and Dense decoding \and DINOv3 \and Histopathology.}
\end{abstract}
\section{Introduction} \label{sec:intro}

Accurate segmentation of tumor bulk in Whole-Slide Images (WSIs) is fundamental for quantifying the Tumor Microenvironment (TME). Quantitative metrics derived from these regions, particularly the Tumor-Stroma Ratio (TSR) and the spatial distribution of Tumor-Infiltrating Lymphocytes (TILs), serve as independent prognostic predictors \cite{salgado2015evaluation,graham2024conic}. Unlike anatomical segmentation in radiology, delineating tumor nests in histopathology presents unique challenges due to inherent biological ambiguity and interpatient heterogeneity. Aggressive tumors exhibit invasive growth patterns where malignant cells disperse into adjacent stroma, forming jagged, discontinuous interfaces that are difficult to distinguish from reactive tissue.

Historically, automated segmentation has relied on task-specific Convolutional Neural Networks (CNNs), particularly U-Net and its nested variants \cite{ronneberger2015u,zhou2018unet++}. Recently, the field is shifting toward adapting large-scale Vision Foundation Models (VFMs), ranging from domain-specific models trained on massive histology archives \cite{chen2024towards} to general-purpose vision experts. Self-supervised encoders, such as DINOv3 \cite{simeoni2025dinov3}, produce dense, high-fidelity representations that have shown promise in fine-grained tasks like nuclei detection \cite{xu2025tissue} and mitotic figure classification \cite{balezo2025efficient}. These developments suggest that DINOv3 possesses the requisite textural sensitivity for tissue-level segmentation. However, current adaptations often pair these powerful frozen encoders with simplistic, linear upsampling decoders \cite{gao2025dino}. We argue that such architectures suffer from a critical "capacity mismatch": the lightweight decoder lacks the structural complexity to reconstruct the intricate, pixel-perfect boundaries of invasive margins from the encoder's high-level semantics, frequently resulting in over-smoothed predictions.

To bridge this gap, we propose \textbf{Dino-NestedUNet}, a hybrid framework that integrates the robust semantic representation of DINOv3 with the structural refinement capabilities of a Nested Dense Decoder (UNet++). Instead of a sparse upsampling path, our architecture establishes a dense grid of intermediate nodes to enforce continuous feature recalibration across multiple resolutions. This design effectively acts as a structural calibrator, progressively fusing deep semantics with low-level textural cues to recover biologically plausible boundaries from coarse region-level supervision.

Our contributions are three-fold: (1) we introduce Dino-NestedUNet, a novel architecture designed to bridge the gap between powerful foundation encoders and simple decoders in histopathology segmentation; (2) by leveraging a nested decoding strategy, we show that the model can resolve infiltrative tumor margin ambiguity and achieve precise boundary recovery from coarse, region-level supervision; and (3) through comprehensive experiments on an internal OSU cohort, a multi-center CHTN dataset, and the CAMELYON16 benchmark, we demonstrate that Dino-NestedUNet significantly surpasses state-of-the-art foundation model adaptations in both predictive performance and cross-dataset generalization.

\section{Methodology}

\begin{figure*}[t]
	\centering
	\includegraphics[width=1\textwidth]{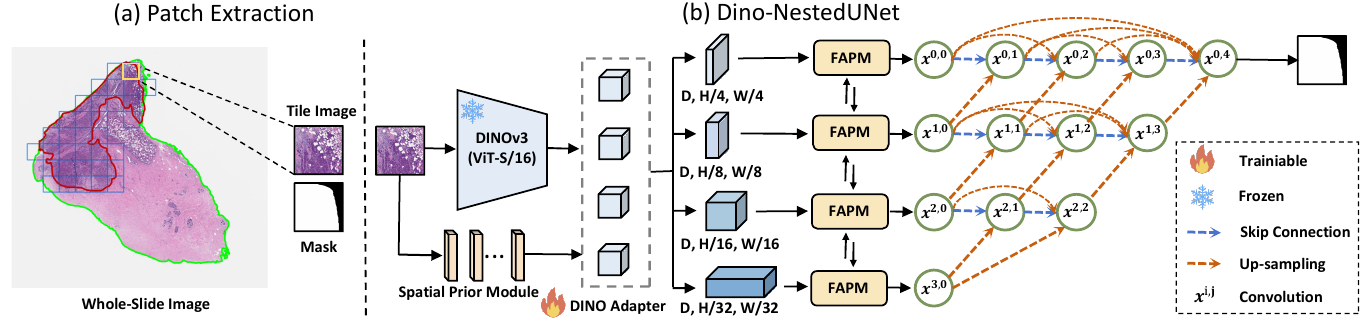} 
	\caption{\textbf{Overview of the Dino-NestedUNet framework.} (a) Annotation-guided patch extraction process. (b) The proposed architecture couples a frozen DINOv3 encoder with a Fidelity-Aware Projection Module (FAPM) and a Nested Dense Decoder. The decoder employs dense skip connections to recursively aggregate multi-scale features for precise boundary reconstruction.}
	\label{overview}
\end{figure*}

\subsection{Overview}
The proposed \textbf{Dino-NestedUNet} framework (Fig. \ref{overview}) is a hybrid segmentation architecture designed to bridge the gap between foundation model semantics and fine-grained boundary reconstruction. The workflow consists of three integrated stages: (1) \textbf{Feature Extraction}, using a frozen DINOv3 backbone coupled with a dual-branch Adapter to extract spatially-enriched semantic features; (2) \textbf{Feature Projection}, employing a Fidelity-Aware Projection Module (FAPM) \cite{gao2025dino} to modulate and project these high-dimensional features into decoder-compatible spaces without information loss; and (3) \textbf{Nested Reconstruction}, utilizing a dense residual decoder to recursively fuse semantic context with textural details for precise margin delineation.

\subsection{DINOv3 Encoder and Adapter}
We leverage DINOv3 (ViT-S/16) \cite{simeoni2025dinov3} as the frozen feature extractor ($E_{\text{Dino}}$). To overcome the lack of intrinsic spatial hierarchy in standard ViT outputs, we adopt a dual-branch adapter architecture.
A \textbf{Spatial Prior Module (SPM)} processes the input image $X$ to capture low-level geometric information, generating a hierarchy of multi-scale feature maps $\{C_i\}_{i=1}^{N}$. In parallel, the frozen DINOv3 backbone extracts high-level semantic features $\{F_{\text{vit}, i}\}_{i=1}^{N}$.
These two streams are fused via a series of Interaction Blocks ($\text{IB}_i$). In each block, the spatial features from the previous stage ($C'_{i-1}$) serve as queries ($Q$) to sample contextual information from the DINOv3 semantic features ($K, V$) via deformable cross-attention ($\mathcal{A}_{\text{de}}$):
\begin{equation}
    C'_{i} = \text{IB}_i(C'_{i-1}, F_{\text{vit}, i}) = \mathcal{A}_{\text{de}}(Q=C'_{i-1}, K=F_{\text{vit}, i}, V=F_{\text{vit}, i})
\end{equation}
This iterative process yields a set of spatially-precise and semantically-rich features $\{C'_i\}_{i=1}^{N}$ at refined scales ($1/4, 1/8, 1/16, 1/32$).

\subsection{Fidelity-Aware Projection Module (FAPM)}
The enriched features $C'_i$ reside in a high-dimensional space incompatible with standard decoders. To reduce dimensionality while preserving fine-grained details, we minimize information loss using the FAPM. For each scale $i$, the feature $C'_i$ is decomposed into a shared context $Z_{ctx,i}$ and scale-specific details $Z_{sp,i}$ via orthogonal $1\times1$ convolutions ($W_{ctx}$, $W_{sp}$):
\begin{equation}
    Z_{ctx,i} = W_{ctx}^T C'_i, \quad Z_{sp,i} = W_{sp,i}^T C'_i
\end{equation}
The shared context $Z_{ctx,i}$ is then used to dynamically modulate the specific details. A small generator $\mathcal{G}_i$ produces affine parameters $(\gamma_i, \beta_i) = \mathcal{G}_i(Z_{ctx,i})$, which recalibrate $Z_{sp,i}$:
\begin{equation}
    Z_{mod,i} = \gamma_i \odot Z_{sp,i} + \beta_i
\end{equation}
Finally, the modulated features undergo spatial refinement via depthwise separable convolutions ($\mathcal{C}_{\text{dwsep}}$) and Squeeze-and-Excitation (SE) recalibration. A residual connection ensures signal fidelity, yielding the final decoder inputs $S_i$:
\begin{equation}
    S_i = \text{SE}(\mathcal{C}_{\text{dwsep}}(Z_{mod,i})) + \mathcal{P}_i(Z_{mod,i})
\end{equation}
where $\mathcal{P}_i$ represents a projection shortcut. These high-fidelity features $S = \{S_i\}_{i=1}^{N}$ form the dense inputs to our Nested Decoder.

\subsection{Nested Dense Decoder}
Standard U-Net decoders rely on sparse skip connections, often failing to recover fine-grained boundary irregularities. We introduce a Nested Dense Decoder (Fig. \ref{overview}b) to enforce dense feature reuse. The decoder is structured as a grid of convolutional nodes $x^{i,j}$, where $i$ indexes the down-sampling layer and $j$ the dense layer ($j=0$ corresponds to $S_i$).

Unlike linear topologies, each node $x^{i,j}$ aggregates feature maps from \textit{all} preceding nodes at the same resolution level, concatenated with the upsampled feature from the lower level. Formally, for any dense node $j>0$:
\begin{equation}
    x^{i,j} = \mathcal{H}\left(\left[ \mathcal{U}(x^{i+1, j-1}), \underbrace{x^{i,0}, x^{i,1}, \dots, x^{i,j-1}}_{\text{Dense Skip Connections}} \right]\right)
\end{equation}
where $\mathcal{H}(\cdot)$ denotes a VGG-style block (Conv-BN-ReLU), $\mathcal{U}(\cdot)$ is bilinear upsampling, and $[\cdot]$ represents channel-wise concatenation.
This dense connectivity ensures that the high-fidelity textural cues preserved in $S_i$ (i.e., $x^{i,0}$) are re-injected at every reconstruction stage, preventing feature dilution and enabling precise delineation of infiltrating tumor margins.

\subsection{Loss Function}
We optimize the segmentation network with a Dice-based compound loss.
For the \textit{Standard} setting (used by the original Dino-UNet), we use Dice + multi-class Cross-Entropy:
\begin{equation}
\mathcal{L}_{\text{std}} = \mathcal{L}_{\text{CE}}(Y, \hat{Y}) + \mathcal{L}_{\text{Dice}}(Y, \hat{Y}).
\end{equation}
For fair comparison with UNet++ and our decoder, we also report a \textit{BCE} variant trained with Dice + Binary Cross-Entropy:
\begin{equation}
\mathcal{L}_{\text{bce}} = \mathcal{L}_{\text{BCE}}(Y, \hat{Y}) + \mathcal{L}_{\text{Dice}}(Y, \hat{Y}).
\end{equation}

\section{Experiments}

 \subsection{Datasets}

\begin{table}[t]
\centering
\caption{Patch-level statistics for the three training cohorts with stratified train/val/test splits (7:1:2). TIGER (93 WSIs, 1,270 patches) is reserved exclusively for zero-shot generalization evaluation.}
\label{tab:dataset_stats}
\resizebox{0.6\textwidth}{!}{%
\begin{tabular}{lcccc}
\toprule
Dataset & WSIs & Train & Val & Test \\
\midrule
CHTN & 300 & 7,300 & 1,043 & 2,084 \\
OSU & 49 & 6,229 & 890 & 1,780 \\
CAMELYON16 & 111 & 3,356 & 478 & 960 \\
\bottomrule
\end{tabular}%
}
\end{table}

We conducted experiments on three training cohorts and one held-out generalization benchmark, covering diverse cancer types, tissue origins, and acquisition centers. All patches were resized to $1024\times1024$ pixels for network training and evaluation, and all datasets were split at the slide level to prevent data leakage. After patch extraction, the three training cohorts contained 10,427 patches from CHTN, 8,899 patches from OSU, and 4,794 patches from CAMELYON16. For each cohort, patches were assigned according to the slide-level split with a train/validation/test ratio of 7:1:2.

\noindent\textbf{CHTN.}
The Cooperative Human Tissue Network (CHTN) annotated subset~\cite{chtn} comprises 300 WSIs spanning 14 cancer types with pixel-level tumor annotations, digitized at $20\times$ magnification (0.5\,$\mu$m/px). This cohort serves as the primary training set, offering substantial histologic diversity across tumor morphologies. After slide-level splitting, CHTN yielded 7,300 training patches, 1,043 validation patches, and 2,084 test patches.

\noindent\textbf{OSU.}
An institutional cohort of 49 colorectal cancer (CRC) WSIs from The Ohio State University Wexner Medical Center, digitized at $40\times$ magnification and manually annotated by expert pathologists. This dataset provides an in-house clinical validation benchmark. After slide-level splitting, OSU yielded 6,229 training patches, 890 validation patches, and 1,780 test patches.

\noindent\textbf{CAMELYON16.}
We utilized the 111 tumor-positive H\&E-stained sentinel lymph node WSIs from the CAMELYON16 benchmark~\cite{bejnordi2017diagnostic}, acquired across two centers at $40\times$ magnification (0.243\,$\mu$m/px), covering tumor boundary, pure tumor, and normal tissue regions. After slide-level splitting, CAMELYON16 yielded 3,356 training patches, 478 validation patches, and 960 test patches.

\noindent\textbf{TIGER.}
To assess zero-shot cross-dataset generalization, we evaluated on 93 tumor-positive WSIs from the TIGER WSIBULK dataset~\cite{shephard2022tiager}, acquired at $20\times$ magnification (0.5\,$\mu$m/px), yielding 1,270 patches. No fine-tuning was performed; all evaluations used the model trained exclusively on CHTN.

\subsection{Implementation Details}

The implementation was conducted on a Linux system using Python 3.10 and PyTorch 2.0 with CUDA 11.8, accelerated by an Nvidia A100-PCIE-40GB GPU. Following \cite{gao2025dino}, the DINOv3 backbone parameters were initialized from LVD-1689M pre-trained weights and kept frozen throughout the training process. The remaining trainable modules were initialized using the default PyTorch initialization scheme. For optimization, these modules were trained using Stochastic Gradient Descent (SGD) with a Nesterov momentum of 0.99, an initial learning rate of $1\times10^{-4}$, and weight decay of $3\times10^{-5}$, governed by a poly learning rate schedule. In contrast, the UNet++ \cite{zhou2018unet++} baseline was optimized using Adam \cite{kingma2014adam} with an initial learning rate of $3\times10^{-4}$ and weight decay of $1\times10^{-4}$. All models were trained for 100 epochs with a batch size of 8, with model weights saved only upon achieving a new best Dice score on the validation set.

\subsection{Evaluation Metrics}
We evaluate tumor bulk segmentation using four standard pixel-level metrics: Dice Similarity Coefficient (Dice), Recall, Precision, and Accuracy. Let $TP$, $TN$, $FP$, and $FN$ denote the numbers of true-positive, true-negative, false-positive, and false-negative pixels, respectively, computed by comparing the predicted mask $\hat{Y}$ with the ground-truth mask $Y$. The metrics are defined as:
\begin{equation}
\mathrm{Dice} = \frac{2TP}{2TP + FP + FN},
\end{equation}
\begin{equation}
\mathrm{Recall} = \frac{TP}{TP + FN},
\end{equation}
\begin{equation}
\mathrm{Precision} = \frac{TP}{TP + FP},
\end{equation}
\begin{equation}
\mathrm{Accuracy} = \frac{TP + TN}{TP + TN + FP + FN}.
\end{equation}
Dice measures overlap between prediction and ground truth, Recall reflects the ability to capture tumor pixels, Precision quantifies the reliability of predicted tumor regions, and Accuracy summarizes overall pixel-wise correctness.

\section{Results and Discussion}

\begin{table}[htb]
\centering
\caption{Quantitative segmentation performance (mean $\pm$ std) on CHTN, OSU, and Camelyon16 datasets. We compare our proposed method against the classical Nested U-Net (UNet++) baseline \cite{zhou2018unet++} and the standard Dino U-Net \cite{gao2025dino}. Note that "Standard" denotes the original Dino U-Net trained with Dice+CrossEntropy loss, while "Standard + BCE" uses BCE+Dice loss to ensure fair comparison with UNet++ and our method. \textbf{Bold} indicates the best performance, and \underline{underline} indicates the second-best performance.}
\label{tab:full_seg_results}
\small
\setlength{\tabcolsep}{6pt}  
\resizebox{\linewidth}{!}{%
\begin{tabular}{@{}llcccc@{}}
\toprule
\textbf{Dataset} & \textbf{Model} & \textbf{Dice} & \textbf{Recall} & \textbf{Precision} & \textbf{Accuracy} \\
\midrule
\multirow{4}{*}{\textbf{CHTN}} 
& UNet++ \cite{zhou2018unet++} & \underline{0.8995 $\pm$ 0.1655} & \underline{0.9373 $\pm$ 0.1110} & \underline{0.8974 $\pm$ 0.1783} & \underline{0.9224 $\pm$ 0.0862} \\
& Dino-UNet (Standard) \cite{gao2025dino} & 0.8664 $\pm$ 0.2207 & 0.9067 $\pm$ 0.2107 & 0.8479 $\pm$ 0.2454 & 0.8715 $\pm$ 0.1783 \\
& Dino-UNet (BCE) & 0.8934 $\pm$ 0.2129 & 0.9134 $\pm$ 0.2094 & 0.8828 $\pm$ 0.2236 & 0.9017 $\pm$ 0.1801 \\
& \textbf{Dino-NestedUNet (Ours)} & \textbf{0.9455 $\pm$ 0.1241} & \textbf{0.9708 $\pm$ 0.0920} & \textbf{0.9415 $\pm$ 0.1405} & \textbf{0.9407 $\pm$ 0.1256} 
 \\
\midrule
\multirow{4}{*}{\textbf{OSU}} 
& UNet++ \cite{zhou2018unet++} & 0.9445 $\pm$ 0.0693 & 0.9472 $\pm$ 0.0816 & \textbf{0.9489 $\pm$ 0.0794} & 0.9408 $\pm$ 0.0672 \\
& Dino-UNet (Standard) \cite{gao2025dino} & 0.9363 $\pm$ 0.0875 & 0.9450 $\pm$ 0.0918 & 0.9401 $\pm$ 0.1061 & 0.9295 $\pm$ 0.0938 \\
& Dino-UNet (BCE) & \textbf{0.9511 $\pm$ 0.0728} & \textbf{0.9660 $\pm$ 0.0725} & 0.9442 $\pm$ 0.0902 & \underline{0.9468 $\pm$ 0.0753} \\
& \textbf{Dino-NestedUNet (Ours)} & \underline{0.9505 $\pm$ 0.0719} & \underline{0.9633 $\pm$ 0.0752} & \underline{0.9451 $\pm$ 0.0888} & \textbf{0.9470 $\pm$ 0.0711} \\
\midrule
\multirow{4}{*}{\textbf{Camelyon16}} 
& UNet++ \cite{zhou2018unet++} 
& 0.8204 $\pm$ 0.2552 
& 0.8573 $\pm$ 0.1896 
& 0.7866 $\pm$ 0.2541 
& 0.9102 $\pm$ 0.1215 \\
& Dino-UNet (Standard) \cite{gao2025dino} 
& 0.7642 $\pm$ 0.2822 
& 0.7873 $\pm$ 0.2700 
& \underline{0.8412 $\pm$ 0.2629} 
& 0.9042 $\pm$ 0.1125 \\
& Dino-UNet (BCE) 
& \underline{0.8254 $\pm$ 0.2475} 
& \underline{0.9043 $\pm$ 0.1381} 
& 0.8287 $\pm$ 0.2636 
& \underline{0.9326 $\pm$ 0.0956} \\
& \textbf{Dino-NestedUNet (Ours)} 
& \textbf{0.8462 $\pm$ 0.2288} 
& \textbf{0.9088 $\pm$ 0.1434} 
& \textbf{0.8497 $\pm$ 0.2398} 
& \textbf{0.9383 $\pm$ 0.0921} \\
\bottomrule
\end{tabular}
}
\end{table}

\begin{table}[t]
\centering
\caption{Cross-dataset generalization of Dino-NestedUNet trained only on CHTN and directly evaluated on unseen external cohorts without fine-tuning.}
\label{tab:cross_dataset}
\small
\begin{tabular}{lccc}
\toprule
\textbf{Training Set} & \textbf{External Test Set} & \textbf{mDice} & \textbf{mIoU} \\
\midrule
CHTN & TIGER WSIBULK (n=93) & 0.8176 & 0.7773 \\
CHTN & OSU CRC (n=49) & 0.8643 & 0.8168 \\
\bottomrule
\end{tabular}
\end{table}

\begin{figure*}[t!]
  \centering
  \includegraphics[width=0.85\linewidth]{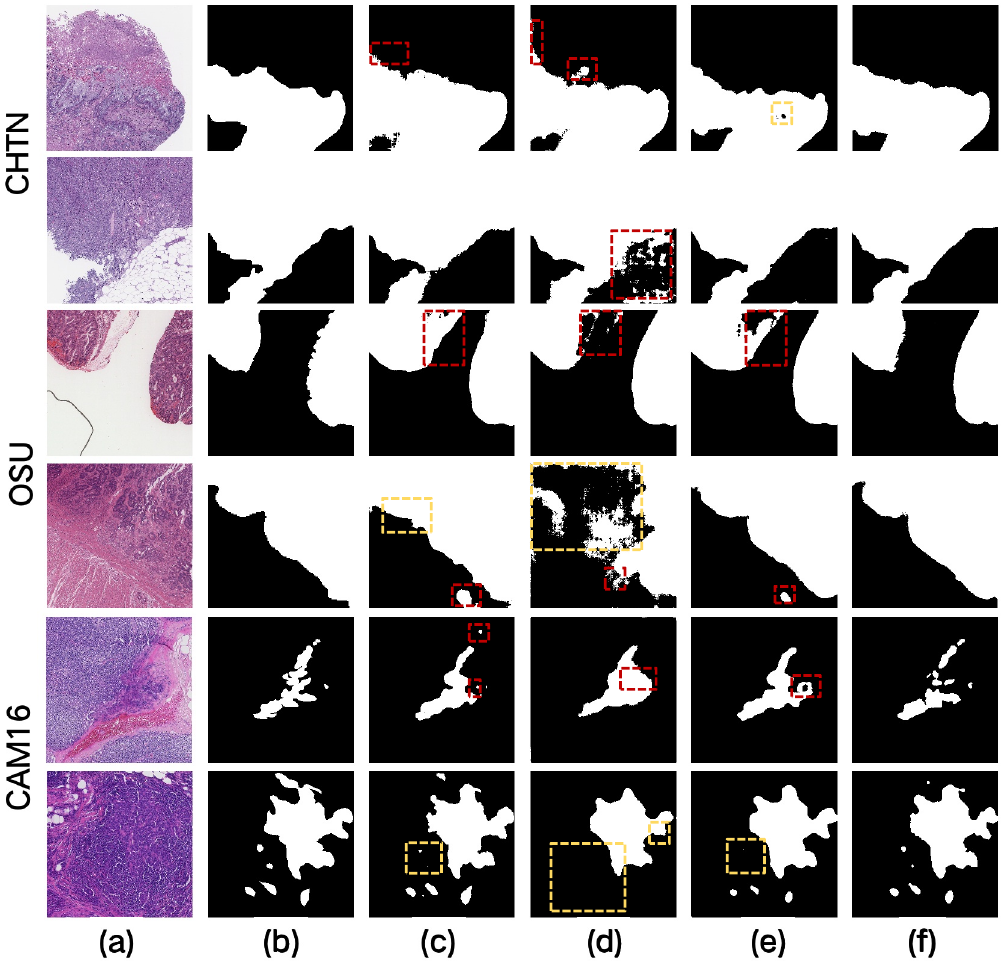}
  
  \caption{\textbf{Comparative segmentation results on CHTN (rows 1--2), OSU (rows 3--4), and CAMELYON16 (rows 5--6), highlighting infiltrative tumor boundaries.} Columns: (a) Input patch, (b) Ground truth, (c) UNet++ \cite{zhou2018unet++}, (d) Dino-UNet (Standard) \cite{gao2025dino}, (e) Dino-UNet (BCE), and (f) Ours (Dino-NestedUNet). \textbf{Red} and \textbf{Yellow} dashed boxes indicate false positives and false negatives, respectively. Compared with prior methods, our model suppresses artifacts and more accurately follows complex tumor boundaries.}
  \label{fig:qualitative_results}
\end{figure*}

\begin{figure*}[t!]
  \centering
  \includegraphics[width=0.7\linewidth]{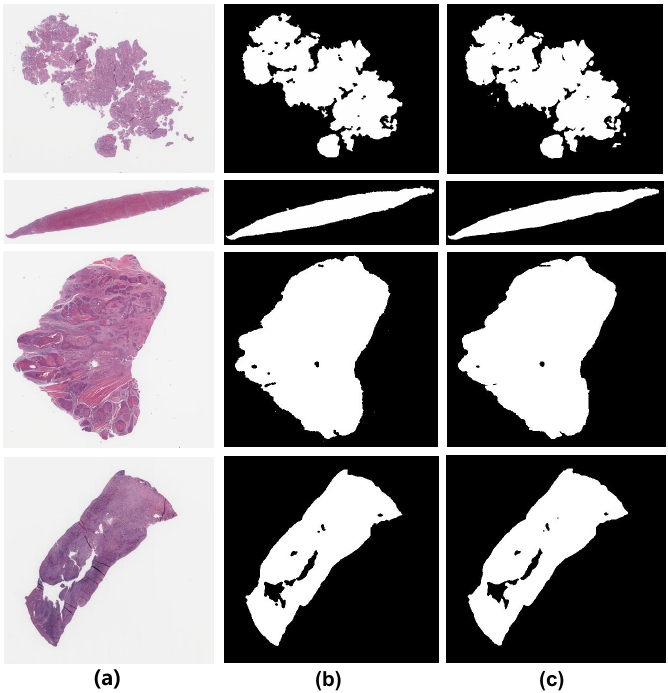}
  \caption{\textbf{Representative whole-slide segmentation results on the CHTN cohort.}
  Representative CHTN whole-slide images showing (a) the original WSI (Hematoxylin \& Eosin stain); (b--c) segmentation masks (white = tumor, black = non-tumor background): (b) ground truth and (c) ours (Dino-NestedUNet).}
  \label{fig:wsi_results}
\end{figure*}

\begin{figure*}[t!]
  \centering
  \includegraphics[width=0.95\linewidth]{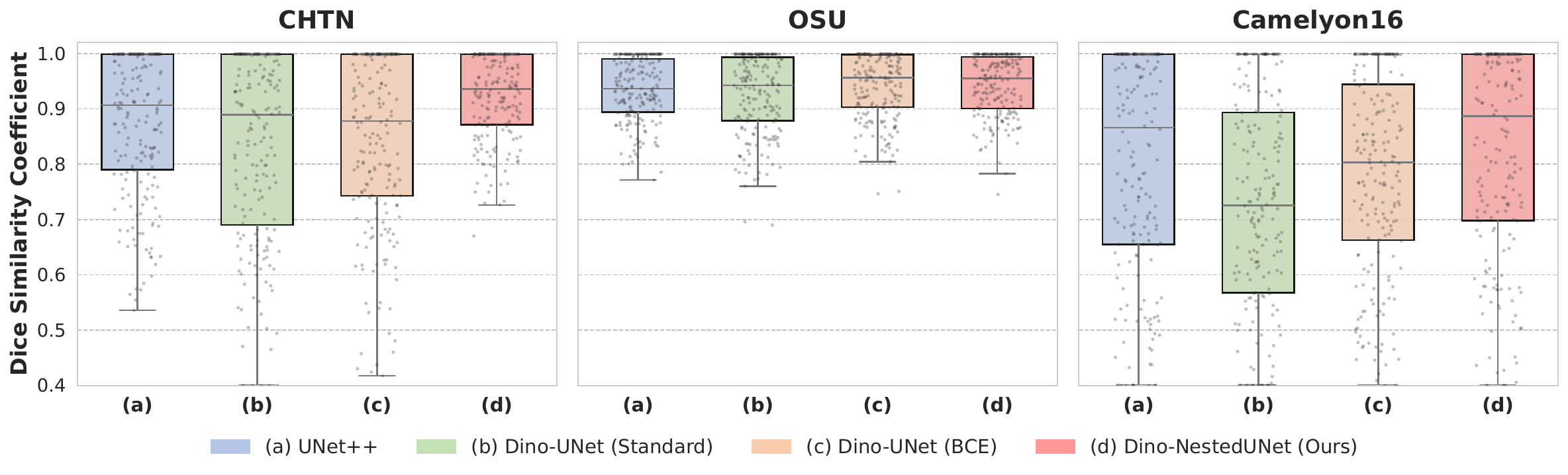}
  \caption{\textbf{Distribution of Dice Similarity Coefficients.} Boxplots illustrating the performance spread of (a) UNet++, (b) Standard Dino-UNet, (c) Dino-UNet (BCE), and (d) Dino-NestedUNet across the three testing cohorts.}
  \label{fig:boxplot}
\end{figure*}

\textbf{Quantitative Analysis.} Table~\ref{tab:full_seg_results} summarizes quantitative performance on CHTN, OSU, and CAMELYON16. Dino-NestedUNet achieves the strongest overall results, with clear gains on the more challenging cohorts. On CHTN, our method improves Dice from 0.8995 (UNet++) to 0.9455 and consistently increases Recall, Precision, and Accuracy, indicating both better tumor coverage and fewer false positives. Notably, the variance is also reduced (Dice std: 0.1241 vs 0.1655/0.2207), suggesting more stable segmentation across heterogeneous slides. On CAMELYON16, Dino-NestedUNet also delivers the best performance across all reported metrics (Dice 0.8462, Accuracy 0.9383), outperforming the strongest baseline (Dino-UNet with BCE) and demonstrating improved generalization under domain shift. On OSU, Dino-NestedUNet remains competitive with the best-performing baseline. While Dino-UNet (BCE) attains a marginally higher Dice and Recall, our method yields the highest Accuracy (0.9470) and comparable Precision, reflecting a favorable balance between sensitivity and over-segmentation control. Overall, these results indicate that dense decoding can better translate foundation-model features into boundary-faithful tumor masks, particularly when appearance varies across cohorts.

\textbf{Cross-dataset generalization.}
To assess whether Dino-NestedUNet learns transferable tumor-bulk representations rather than dataset-specific appearance patterns, we performed zero-shot external validation. The model was trained only on CHTN and directly evaluated on TIGER WSIBULK and OSU CRC without fine-tuning. The CHTN-trained model achieved an mDice of 0.8176 and an mIoU of 0.7773 on TIGER WSIBULK, and an mDice of 0.8643 and an mIoU of 0.8168 on OSU CRC. These results indicate that the proposed architecture maintains strong segmentation performance under external domain shift, supporting its ability to generalize across unseen cohorts.

\textbf{Qualitative Analysis.} Qualitative comparisons in Fig.~\ref{fig:qualitative_results} further support these findings. UNet++ and Dino-UNet variants exhibit typical failure modes in histopathology tumor bulk segmentation, including spurious activations in normal regions (false positives, red boxes) and missed thin or ambiguous tumor extensions (false negatives, yellow boxes). Dino-UNet (Standard) is prone to noisy foreground fragments in challenging regions, while the BCE variant reduces some false positives but may still miss small boundary structures. In contrast, Dino-NestedUNet produces cleaner masks with tighter boundary adherence and fewer isolated artifacts across all three datasets, consistent with the improved Precision/Accuracy and reduced variance observed quantitatively. This pattern also extends to the WSI level: as shown in Fig.~\ref{fig:wsi_results}, the predicted masks closely align with the ground truth and accurately capture the overall tumor extent across diverse CHTN specimens.

\textbf{Distribution and Stability Analysis.} To further investigate the robustness of our model beyond mean metrics, we visualize the distribution of Dice scores in Fig. \ref{fig:boxplot}. On the multi-center CHTN and public CAMELYON16 datasets, our Dino-NestedUNet (Column d) exhibits a visibly superior distribution profile. The median Dice scores are consistently higher than those of the UNet++ (a) and Standard Dino-UNet (b) baselines. More importantly, the box plots for our method are more compact with shorter lower whiskers, indicating a significantly higher performance floor. This suggests that the dense nested decoding strategy effectively mitigates "corner case" failures where standard models collapse (e.g., Dice $<0.6$), ensuring reliable segmentation even on difficult samples. On the internal OSU dataset, while the mean performance is competitive with the BCE variant (c), our model maintains a tighter interquartile range, reflecting high consistency across the cohort. In contrast, the Standard Dino-UNet (b) shows a dispersed distribution with numerous low-scoring outliers, confirming that the vanilla linear decoder struggles to maintain stability when handling complex histological variations.

\section{Conclusion}
We presented Dino-NestedUNet for pathology tumor bulk segmentation and showed that improving the decoder is an effective way to better exploit strong frozen vision encoders for boundary-sensitive prediction. Across three cohorts, the model produced cleaner and more boundary-faithful tumor masks than existing baselines, with the clearest gains observed on the more heterogeneous evaluation settings. These findings suggest that dense nested decoding is a practical design choice for adapting foundation-model features to pathology segmentation. Future work will extend this framework beyond patch-level inference and further evaluate its robustness across broader datasets and tumor types.

\section{Disclosure of Interests}

The authors have no competing interests in the paper.

\bibliographystyle{splncs04}
\bibliography{references}

@article{gao2025dino,
  title={Dino u-net: Exploiting high-fidelity dense features from foundation models for medical image segmentation},
  author={Gao, Yifan and Li, Haoyue and Yuan, Feng and Wang, Xiaosong and Gao, Xin},
  journal={arXiv preprint arXiv:2508.20909},
  year={2025}
}

@article{bejnordi2017diagnostic,
  title={Diagnostic assessment of deep learning algorithms for detection of lymph node metastases in women with breast cancer},
  author={Bejnordi, Babak Ehteshami and Veta, Mitko and Van Diest, Paul Johannes and Van Ginneken, Bram and Karssemeijer, Nico and Litjens, Geert and Van Der Laak, Jeroen AWM and Hermsen, Meyke and Manson, Quirine F and Balkenhol, Maschenka and others},
  journal={Jama},
  volume={318},
  number={22},
  pages={2199--2210},
  year={2017},
  publisher={American Medical Association}
}

@article{shephard2022tiager,
  title={Tiager: Tumor-infiltrating lymphocyte scoring in breast cancer for the tiger challenge},
  author={Shephard, Adam and Jahanifar, Mostafa and Wang, Ruoyu and Dawood, Muhammad and Graham, Simon and Sidlauskas, Kastytis and Khurram, Syed Ali and Rajpoot, Nasir and Raza, Shan E Ahmed},
  journal={arXiv preprint arXiv:2206.11943},
  year={2022}
}

@inproceedings{zhou2018unet++,
  title={Unet++: A nested u-net architecture for medical image segmentation},
  author={Zhou, Zongwei and Rahman Siddiquee, Md Mahfuzur and Tajbakhsh, Nima and Liang, Jianming},
  booktitle={International workshop on deep learning in medical image analysis},
  pages={3--11},
  year={2018},
  organization={Springer}
}

@article{simeoni2025dinov3,
  title={Dinov3},
  author={Sim{\'e}oni, Oriane and Vo, Huy V and Seitzer, Maximilian and Baldassarre, Federico and Oquab, Maxime and Jose, Cijo and Khalidov, Vasil and Szafraniec, Marc and Yi, Seungeun and Ramamonjisoa, Micha{\"e}l and others},
  journal={arXiv preprint arXiv:2508.10104},
  year={2025}
}

@inproceedings{ronneberger2015u,
  title={U-net: Convolutional networks for biomedical image segmentation},
  author={Ronneberger, Olaf and Fischer, Philipp and Brox, Thomas},
  booktitle={International Conference on Medical image computing and computer-assisted intervention},
  pages={234--241},
  year={2015},
  organization={Springer}
}

@article{xu2025tissue,
  title={Tissue Aware Nuclei Detection and Classification Model for Histopathology Images},
  author={Xu, Kesi and Chiou, Eleni and Varamesh, Ali and Acqualagna, Laura and Rajpoot, Nasir},
  journal={arXiv preprint arXiv:2511.13615},
  year={2025}
}

@article{chen2024towards,
  title={Towards a general-purpose foundation model for computational pathology},
  author={Chen, Richard J and Ding, Tong and Lu, Ming Y and Williamson, Drew FK and Jaume, Guillaume and Song, Andrew H and Chen, Bowen and Zhang, Andrew and Shao, Daniel and Shaban, Muhammad and others},
  journal={Nature medicine},
  volume={30},
  number={3},
  pages={850--862},
  year={2024},
  publisher={Nature Publishing Group US New York}
}

@article{balezo2025efficient,
  title={Efficient Fine-Tuning of DINOv3 Pretrained on Natural Images for Atypical Mitotic Figure Classification (MIDOG 2025 Task 2 Winner)},
  author={Balezo, Guillaume and Feki, Hana and Bourgade, Rapha{\"e}l and Monnier, Lily and Blons, Matthieu and Blondel, Alice and Decenci{\`e}re, Etienne and Planas, Albert Pla and Walter, Thomas},
  journal={arXiv preprint arXiv:2508.21041},
  year={2025}
}

@misc{chtn,
  title={Cooperative Human Tissue Network ({CHTN})},
  author={{National Cancer Institute}},
  howpublished={\url{https://www.chtn.org}},
  year={2024}
}

@article{salgado2015evaluation,
  title={The evaluation of tumor-infiltrating lymphocytes (TILs) in breast cancer: recommendations by an International TILs Working Group 2014},
  author={Salgado, Roberto and Denkert, Carsten and Demaria, Sandra and Sirtaine, Nicolas and Klauschen, Frederick and Pruneri, Giancarlo and Wienert, S and Van den Eynden, Gert and Baehner, Frederick L and P{\'e}nault-Llorca, Frederique and others},
  journal={Annals of oncology},
  volume={26},
  number={2},
  pages={259--271},
  year={2015},
  publisher={Elsevier}
}

@article{graham2024conic,
  title={CoNIC Challenge: Pushing the frontiers of nuclear detection, segmentation, classification and counting},
  author={Graham, Simon and Vu, Quoc Dang and Jahanifar, Mostafa and Weigert, Martin and Schmidt, Uwe and Zhang, Wenhua and Zhang, Jun and Yang, Sen and Xiang, Jinxi and Wang, Xiyue and others},
  journal={Medical image analysis},
  volume={92},
  pages={103047},
  year={2024},
  publisher={Elsevier}
}

@article{kingma2014adam,
  title={Adam: A method for stochastic optimization},
  author={Kingma, Diederik P},
  journal={arXiv preprint arXiv:1412.6980},
  year={2014}
}

\end{document}